\documentclass{article}

\usepackage[preprint]{neurips_2026}

\usepackage[dvipsnames]{xcolor}
\definecolor{linkColor}{RGB}{38,76,115}
\definecolor{citecolor}{RGB}{0,105,92}
\definecolor{urlColor}{RGB}{74,85,104}
\usepackage[colorlinks=true,linkcolor=linkColor,citecolor=citecolor,filecolor=linkColor,urlcolor=linkColor]{hyperref}

\usepackage[utf8]{inputenc} 
\usepackage[T1]{fontenc}    
\usepackage{hyperref}       
\usepackage{url}            
\usepackage{booktabs}       
\usepackage{amsfonts}       
\usepackage{nicefrac}       
\usepackage{microtype}      
\usepackage{xcolor}         

\usepackage{graphicx}
\usepackage{tabularx}
\usepackage{multirow}
\usepackage{booktabs}
\usepackage{bbding}
\usepackage{caption}
\usepackage{pifont}
\usepackage{stfloats}
\usepackage{bm}
\usepackage{wrapfig}
\usepackage{colortbl}
\usepackage{enumitem}
\usepackage{indentfirst}
\usepackage{algorithm}
\usepackage{algpseudocode}
\usepackage{amsmath}
\usepackage{wrapfig}
\usepackage{booktabs}

\usepackage{fancyhdr} 
\usepackage[most]{tcolorbox} 

\usepackage{caption}
\usepackage{subfigure}

\newcommand{\xmark}{\textcolor{red}{\ding{55}}}
\newcommand{\cmark}{\textcolor{green!60!black}{\ding{51}}}

\usepackage{xspace}
\newcommand{\modelname}{\texttt{MiniMax-H3}\xspace}

\usepackage{duckuments}

\title{Can \modelname Reason About the Physical World? \\ An Evaluation of Omni-Modal Generative Model}

\vspace{-5pt}
\author{
  \vspace{-25pt}\\
  \textbf{
  Haoyu Zhao$^{1,*,\dagger}$,\quad
  Zihao Zhao$^{1,*}$,\quad
  Tianyu Deng$^{1,*}$,\quad
  Ziqin Xu$^{1,*}$,\quad
  Zihao Zhang$^{2}$,
  }\\[0pt]
  \textbf{
  Xudong Wang$^{1}$,\quad
  Jinxiang Guo$^{1}$,\quad
  Chen Gao$^{1}$,\quad
  Ziyi Ye$^{2}$,\quad
  Yeying Jin$^{3,\ddagger}$,\quad
  }\\[0pt]
  \textbf{
  Jiaxi Gu$^{3,\ddagger}$,\quad
  Zuxuan Wu$^{2}$,\quad
  Shuicheng Yan$^{1}$
  }
  \vspace{3pt}\\
  $^1$National University of Singapore
  \quad
  $^2$Fudan University
  \quad
  $^3$Tencent
  \vspace{-4pt}
}

\begin{document}

\maketitle

\begingroup
\renewcommand{\thefootnote}{\fnsymbol{footnote}}

\footnotetext[1]{
Equal contribution.
}

\footnotetext[2]{
Project lead.
}

\footnotetext[3]{
Corresponding authors.
}

\endgroup

\begin{abstract}
Recent Omni-Modal Generative Models (Omni-Models) have advanced content generation toward unified modeling of text, images, video, and audio. \modelname exemplifies this transition by combining multimodal context understanding with joint audio-visual generation in a shared latent framework. Its unified architecture raises a fundamental question: Can multimodal alignment improve the model’s world reasoning, and what new evaluation paradigms do omni-modal inputs enable? To investigate this question, this work introduces a comprehensive evaluation framework organized around four complementary dimensions of physical world reasoning.
Unlike existing evaluation frameworks for video generation and world models, which are often constrained by limited input modalities and evaluation settings where prompts closely match the target video content, our evaluation is specifically designed to exploit the multimodal inputs of Omni-Model.
We construct a diverse set of novel tasks that require models to integrate complementary information across modalities. Specifically, we consider four scenarios, including implicit prompts paired with multiple frames, audio-image, prefix-videos, and audio-video inputs. Every single modality provides only partial evidence about the underlying event, requiring the model to jointly reason over the complementary semantic cues to infer latent event states and future dynamics.
Across 517 evaluation instances, \modelname achieves an overall success rate of 41.97\%. Video-based Decision Reasoning yields the highest success rate at 56.00\%, while Audio-based Disambiguation Reasoning is the weakest, reaching only 27.40\%.
These results indicate that effective multimodal integration remains key to fully exploiting the benefits of diverse input modalities.
The project is available at \href{https://github.com/gulucaptain/MiniMax-H3-Reason}{https://github.com/gulucaptain/MiniMax-H3-Reason}.
\end{abstract}

\section{Introduction}
\label{sec:intro}

Omni-modal generative models (Omni-Models) have recently emerged as a new class of generative systems capable of processing heterogeneous inputs, including text, images, videos, and audio, within a unified architecture~\citep{liu2025ola,li2025baichuan,yang2025humanomniv2,luo2026next,nvidiaWhatOmni}.
Compared with conventional generative models operating on one or two conditioning modalities, Omni-Model can receive multiple observations of the same underlying event and generate audio-visual content conditioned on their joint context.
Recent models such as \modelname~\citep{minimaxh32026} further combine multimodal context understanding with joint audio-visual generation, making it possible to provide the model with different forms of partial observations and inspect its interpretation through the generated result.
Such a setting offers a natural interface for studying whether generative models can reason over complementary information describing the physical world.

Understanding the reasoning capabilities of Omni-Model is a central issue for their use as general-purpose generative models, and a particularly important question is whether they can infer an underlying event from multimodal evidence that is incomplete when considered separately.
The omni-modal inputs create the possibility of grounding generation in substantially richer observations of the physical world.
However, accepting multiple modalities is not equivalent to reasoning across them. A model may support omni-modal inputs while ignoring non-dominant evidence, failing to establish cross-modal correspondences, or relying primarily on semantic priors from the textual prompt. This distinction motivates the question of our work: \emph{Can multimodal alignment improve Omni-Model’s world reasoning, and what new evaluation paradigms do omni-modal inputs enable?}

\begin{figure}[t!]
    \centering
    \includegraphics[width=1.0\linewidth]{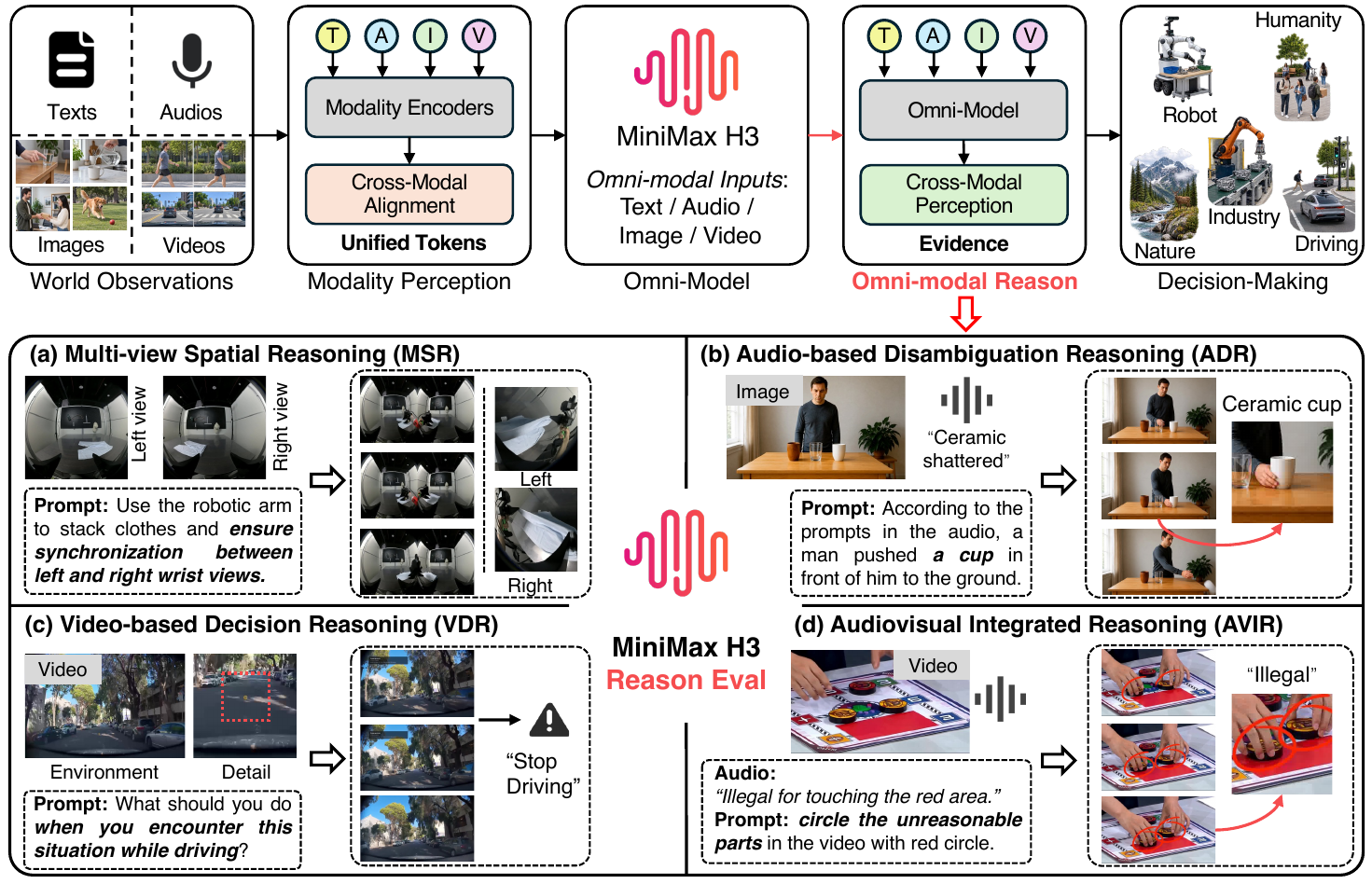}
    \caption{\textbf{Overview of our evaluation for the reasoning capability of \modelname.} First, we illustrate how \modelname progresses from world observation and modality perception to omni-modal model training, where the acquired multimodal knowledge is transformed into task-oriented decision-making through omni-modal reasoning. Second, we construct implicitly paired multimodal data to systematically evaluate it across four complementary dimensions of physical-world reasoning.
}
    \label{fig:intro}
\end{figure}

Existing general benchmarks, including VBench~\citep{huang2024vbench} for video generation and WorldModelBench~\citep{li2026worldmodelbench} for world models, provide limited insight into this question.
In most evaluations, the prompt explicitly describes the expected output, while additional modalities serve as redundant or local conditioning signals. 
Consequently, a model can often produce a plausible result without identifying the relationships among its inputs. Such benchmarks primarily measure whether the model can faithfully render a specified event, but reveal little about whether it can infer an event from distributed multimodal evidence. Put differently, existing benchmarks ask whether a model can generate \emph{what it is told}; we instead ask whether it can determine \emph{what it should generate}.

To this end, we introduce an evaluation framework based on implicit Omni-Model generation. Rather than describing the complete target event in the textual prompt, the key idea is that we deliberately omit critical event semantics and distribute the missing information across multiple input modalities. Each modality provides only partial evidence, and the intended event is recoverable only by aligning and jointly interpreting the observations. The model must therefore identify the relevant evidence, establish its cross-modal relationships, infer the underlying event, and complete that event through video generation. The generated video provides a behavioral readout of this process: its consistency with the multimodal evidence allows us to assess whether the model has recovered the missing semantics, rather than merely followed an explicit generation instruction.

We instantiate this framework through four complementary evaluation settings, shown in Fig.~\ref{fig:intro}. Multi-view Spatial Reasoning tests whether the model can associate entities and spatial relationships across multiple visual observations. Audio-based Disambiguation Reasoning uses acoustic evidence to resolve events that remain ambiguous from visual appearance alone. Video-based Decision Reasoning evaluates whether the model can infer a compatible event continuation from the dynamics observed in a prefix video. Finally, Audiovisual Integrated Reasoning requires the joint interpretation of temporally related visual and acoustic evidence.
These settings examine cross-view association, semantic disambiguation, temporal reasoning, and audiovisual integration within Omni-Model.

We use \modelname as a testbed to examine how reliably generated videos satisfy the requirements supported by their input observations. Our contributions are threefold:

\begin{itemize}
\item We introduce a framework for evaluating physical-world reasoning through video generation, continuation, and editing. Prompts leave task-relevant information unspecified, and outputs are assessed against semantic constraints supported by the input observations.

\item We construct an expert-verified evaluation set of 517 instances across four reasoning scenarios and 29 subcategories, as summarized in Table~\ref{tab:benchmark_comparison}. Each instance pairs input observations with an implicit task prompt and an annotated semantic target, supporting human evaluation against task-specific success criteria.

\item Our quantitative evaluation and qualitative analysis reveal a gap between multimodal input support and reliable task completion in \modelname. Overall success is 41.97\%, with video-based decision reasoning performing best at 56.00\% and audio-based disambiguation performing worst at 27.40\%. These findings expose a gap between supporting multimodal inputs and reliably translating the available evidence into successful task outcomes.
\end{itemize}

\begin{table*}[t!]
\centering
\caption{
\textbf{Comparison with existing evaluation benchmarks.} 
Our evaluation extends beyond explicit text-image conditioning by introducing omni-modal inputs and implicit prompts, enabling the evaluation of multi-scene understanding, physical awareness, and multimodal reasoning.
}
\label{tab:benchmark_comparison}
\resizebox{\textwidth}{!}{
\begin{tabular}{l|ccc|ccc}
\toprule
\textbf{Evaluation Sets}
& \textbf{\# Examples}
& \textbf{Input Modalities}
& \textbf{Prompt Type}
& \textbf{Multi Scene}
& \textbf{Phys. Aaware}
& \textbf{Reasoning} \\
\midrule

VBench~\citep{huang2024vbench}
& 800
& T + I
& Explicit
& \xmark
& \xmark
& \xmark \\

TC-Bench~\citep{feng2024tc}
& 150
& T + I
& Explicit
& \xmark
& \xmark
& \xmark \\

VideoPhy~\citep{bansal2025videophy}
& 688
& T + I
& Explicit
& \xmark
& \cmark
& \xmark \\

WorldModelBench~\citep{li2026worldmodelbench}
& 350
& T + I
& Explicit
& \xmark
& \cmark
& \xmark \\

\midrule

\textbf{Ours}
& 517
& \textbf{T + I + A + V}
& \textbf{Implicit}
& \cmark
& \cmark
& \cmark \\

\bottomrule
\end{tabular}
}
\end{table*}

\section{Related Work}
\label{sec:related}

\paragraph{Video generation evaluation.}
Video generation models focus on perceptual quality and semantic alignment~\citep{huang2024vbench,zhao2024magdiff,zheng2026refocuseraser}, temporal compositionality~\citep{feng2024tc,zhao2026dynamictrl,zhao2026lstd}, and physical faithfulness~\citep{zheng2025vbench,bansal2025videophy,bansal2026videophy,lin2026phyground,zhao2026cameranoise,zhang2026speed}.
Beyond prompt-conditioned synthesis, Physics-IQ~\citep{motamed2026generative} tests physical prediction from observed frames without disclosing future outcomes, while Morpheus~\citep{tragoudaras2025evaluating} evaluates generated dynamics against physical laws.
Evaluation also extends to multimodal settings: AV-Phys Bench~\citep{cui2026joint} examines physical consistency within and across generated audio-video streams. ROVER~\citep{liang2026rover} and OmniVideoBench~\citep{li2026omnivideobench} study cross-modal reasoning through text-image generation and audio-visual question answering, respectively.
Our evaluation connects these directions by assessing whether observations support an inference expressed through video generation.

\paragraph{Reasoning through video generation.}
The zero-shot capabilities of video models~\citep{wiedemer2025video} have motivated benchmarks that evaluate generated sequences as task solutions.
VideoThinkBench~\citep{tong2026thinking}, TiViBench~\citep{chen2025tivibench}, and Gen-ViRe~\citep{liu2025can} probe visual, symbolic, and planning capabilities.
RISE-Video~\citep{liu2026rise} explicitly tests implicit world-rule reasoning, while \citet{zhang2026thinking} examine the gap between causal perception and generated consequences.
Building on these precedents, we focus on how task-relevant information is distributed across inputs.
Our prompts leave the target inference unstated, requiring models to establish cross-view correspondences and infer responses from temporal context.

\paragraph{World-model evaluation.}
As General-Level~\citep{fei2025path} argues that stronger model capabilities bring us closer to human-level AI, several recent works have also sought to evaluate world models.
WorldModelBench~\citep{li2026worldmodelbench} evaluates instruction following and physics adherence in application-driven domains.
WorldScore~\citep{duan2025worldscore} assesses successive scene generation under specified camera trajectories, while WorldMark~\citep{xu2026worldmark} and Omni-WorldBench~\citep{wu2026omni} evaluate control alignment, world consistency, and action-dependent state transitions.
These benchmarks test whether generated environments preserve spatial structure and respond coherently to interactions.
Our framework complements them by examining how observations determine the event or response to generate: spatial tasks require integrating complementary views, and decision tasks require inferring an appropriate response from observed dynamics.
Success is measured by whether the generated outcome satisfies the semantic requirements supported by the input evidence.

\section{Reasoning Evaluation with \modelname}
\label{sec:method}

In this research, we present a systematic evaluation pipeline for Omni-Modes, with a particular focus on assessing the physical-world understanding and reasoning capabilities of \modelname.
Compared with earlier generative models, such as Sora~\citep{liu2024sora}, LTX-Video~\citep{hacohen2024ltx}, or the Wan series~\citep{wan2025wan}, whose conditioning modalities are primarily images and text, \modelname supports compositional inputs across multiple modalities.
By supporting complex multimodal inputs, Omni-Model can integrate complementary cross-modal evidence to perform reliable physical-world reasoning.
In this section, we first compare the supported modality combinations of \modelname with those of existing models. We then formulate physical-world reasoning tasks across four representative scenarios. Finally, we describe the construction of the evaluation data and the human annotation protocol for our evaluation.

\subsection{What Do Omni-Models Enable?}
\label{sec:omgm_capabilities}

\begin{wrapfigure}{r}{0.48\columnwidth}
    \vspace{-0.8em}
    \centering
    \includegraphics[width=1.0\linewidth]{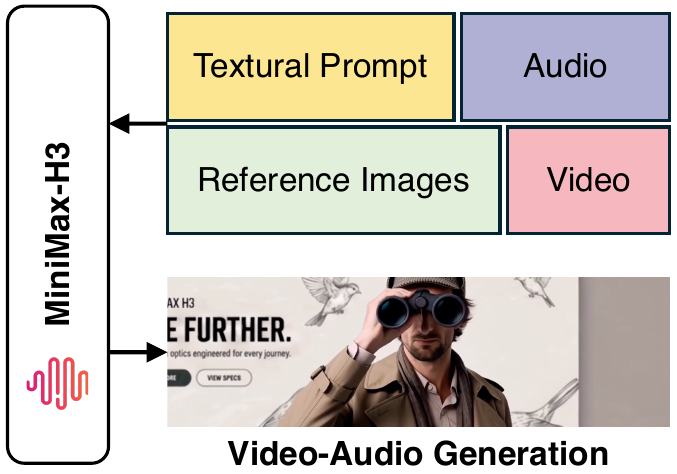}
    \caption{\textbf{Video-audio generation pipeline of \modelname}, which is an open-weight, general-purpose, omni-modal generation model.}
    \label{fig:pipeline}
    \vspace{-0.8em}
\end{wrapfigure}

\noindent
Omni-Model provides a unified interface for conditioning generation on text, images, audio, and video, allowing different modalities to contribute complementary information about the same scene. Images describe visible entities and spatial layouts, video provides motion and state changes, audio offers event cues or spoken constraints, and text specifies the requested operation. This makes it possible to design tasks where part of the target behavior is intentionally left unspecified in the prompt and must instead be inferred from the observations. For example, audio can determine which object in an image should become active, while multiple views can provide the geometry needed to complete a manipulation. The generated video then makes the model's interpretation observable through object motion and state transitions, enabling evaluation based on whether the output is consistent with the available evidence rather than whether it matches a single condition. Our study therefore tests whether the Omni-Model can use the provided evidence to satisfy generation, and answers whether multimodal input supports correct reasoning for the physical world.

\subsection{Evaluation on Physical-World Reasoning Tasks}
\label{sec:reasoning_scenarios}

We evaluate the physical-world reasoning capabilities of \modelname through four tasks: Multi-view Spatial Reasoning (MSR), Audio-based Disambiguation Reasoning (ADR), Video-based Decision Reasoning (VDR), and Audiovisual Integrated Reasoning (AVIR). These tasks probe whether the model can integrate spatial, temporal, and auditory evidence to support inferences expressed through video generation, continuation, or editing. Given a task prompt $q$ and observations $\mathbf{x}$, the model generates an output video:

\begin{equation}
    \hat{Y} \sim p_{\theta}(Y \mid q, \mathbf{x}),
    \label{eq:reasoning_general}
\end{equation}

where $p_{\theta}$ denotes the conditional distribution over output videos induced by \modelname. Text prompts specify the task without explicitly providing the target inference, requiring the model to derive it from the accompanying observations. We assess whether the output video reflects this inference while remaining consistent with the observed scene.

\paragraph{Multi-view Spatial Reasoning (MSR).}
MSR evaluates whether the model can integrate complementary views of a scene to infer its spatial structure (Fig.~\ref{fig:intro} (a)). Given $K$ images $\mathbf{I} = \{I^{(k)}\}_{k=1}^{K}$ captured from different viewpoints, the model generates:

\begin{equation}
    \hat{Y}_{\mathrm{MSR}}
    \sim
    p_{\theta}(Y \mid q, \mathbf{I}).
    \label{eq:msr}
\end{equation}

Instances are constructed so that the target spatial inference depends on evidence distributed across views. The model must establish cross-view correspondences, account for viewpoint changes and occlusions, and infer relationships that are not fully observable from a single image. The generated video is evaluated for consistency with the spatial relationships jointly supported by the input views.

\paragraph{Audio-based Disambiguation Reasoning (ADR).}
ADR evaluates whether audio can resolve ambiguity in a static visual observation (Fig.~\ref{fig:intro} (b)). Given an image $I$ that admits multiple plausible interpretations and an audio input $A$ that provides discriminative evidence, the model generates:

\begin{equation}
    \hat{Y}_{\mathrm{ADR}}
    \sim
    p_{\theta}(Y \mid q, I, A).
    \label{eq:adr}
\end{equation}

The image alone leaves the target interpretation underdetermined, while the audio provides evidence that distinguishes among plausible alternatives.
The model must identify relevant acoustic cues, associate them with the depicted objects or events, and generate a video consistent with the interpretation supported by both modalities.
\textit{For example}, when one of three cups made of different materials falls off a table, the resulting sound provides evidence for identifying which cup fell.
ADR thus assesses whether acoustic evidence informs the model's interpretation of an otherwise ambiguous visual scene.

\paragraph{Video-based Decision Reasoning (VDR).}
VDR evaluates whether the model can infer potential consequences of observed events and generate a continuation that reflects an appropriate response (Fig.~\ref{fig:intro} (c)). Given a prefix video $V_{1:T} = (V_1,\ldots,V_T)$, the model generates:

\begin{equation}
    \hat{Y}_{\mathrm{VDR}}
    \sim
    p_{\theta}(Y \mid q, V_{1:T}),
    \label{eq:vdr}
\end{equation}

where $Y$ denotes the video continuation. The prefix provides temporal evidence about motion, state changes, and interactions, from which the model must infer how an Omni-Model should respond without an explicit action specification in $q$.
\textit{For example}, a ball rolling into the road may indicate that a child could follow, motivating the vehicle to stop before the potential hazard becomes visible. Evaluation focuses on whether the model's behavior in the generated continuation accounts for plausible consequences of the observed events while remaining consistent with the scene dynamics.

\paragraph{Audiovisual Integrated Reasoning (AVIR).}
AVIR evaluates whether the model can integrate video context with auditory evidence or spoken constraints to infer how a video should continue or be revised (Fig.~\ref{fig:intro} (d)). Given a video $V$ and an audio input $A$, the model generates:

\begin{equation}
    \hat{Y}_{\mathrm{AVIR}}
    \sim
    p_{\theta}(Y \mid q, V, A).
    \label{eq:avir}
\end{equation}

The video may be a prefix or a complete sequence, and the audio need not be temporally aligned with it. For video continuation, the model combines the observed dynamics with audio content to infer subsequent events. For video editing, it grounds spoken constraints in the video to identify erroneous content. Whereas ADR focuses on disambiguating a static observation, AVIR requires interpreting audio in the context of a sequence of visual events. This task assesses whether the continuation or revision incorporates the relevant audio content while maintaining consistency with the video.

\begin{figure}[t!]
    \centering
    \includegraphics[width=1.0\linewidth]{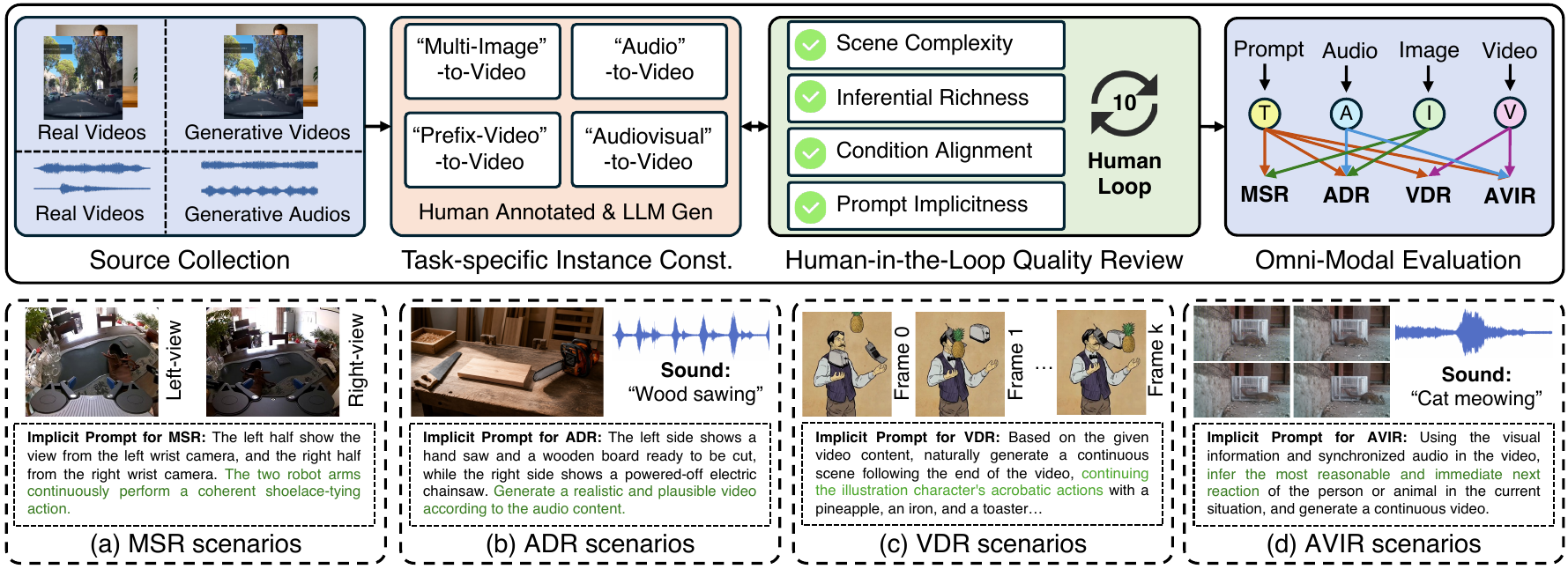}
    \caption{\textbf{Pipeline of evaluation data construction.} We build implicit multimodal condition–prompt pairs in which the information required for correct generation is not explicitly stated in the prompt, but must be inferred from complementary visual and acoustic evidence.}
    \label{fig:daga_pipeline}
\end{figure}

\subsection{Evaluation Data Construction and Human Annotation}
\label{sec:data_construction}

We construct evaluation data through multimodal source collection, task-specific instance construction, and iterative human review, as illustrated in Fig.~\ref{fig:daga_pipeline}. Each instance pairs multimodal observations with an implicit task prompt and an annotated semantic target. The prompt specifies the requested generation operation but leaves task-relevant information to be inferred from the observations. The construction process focuses on whether the available evidence supports an assessable inference whose consequences can be expressed in the output video.

\paragraph{Source Collection.}
We collect real and synthetic visual and acoustic data, including images, videos, audio recordings, and clips produced by generative models, \textit{e.g.,} ChatGPT Voice~\citep{openai2022chatgpt} for audio generation and Seedance 2.0~\citep{seedance2026seedance} for video generation. These sources support variation in scene configurations, object materials, viewpoints, event dynamics, and acoustic cues. Synthetic data additionally allow controlled construction of conditions that are difficult to obtain from existing recordings.
Furthermore, we screen each source for perceptual quality, semantic coherence, and suitability for the intended task. Sources are excluded when visual artifacts or unclear event structure compromise the evidence required for evaluation.

\paragraph{Task-Specific Instance Construction.}
In the test data, we construct each instance by selecting the input observations, defining the output target, and specifying the video generation prompts. The inputs are organized according to the four reasoning scenarios: 1) \textbf{MSR:} Multiple views of the same scene provide complementary evidence for a spatial relation that is not fully specified by any individual view. 2) \textbf{ADR:} A static image admits multiple plausible interpretations, while an accompanying audio clip provides evidence for distinguishing among them. 3) \textbf{VDR:} A video prefix contains motion, state changes, or interactions that support an anticipatory response to be expressed in the continuation. 4) \textbf{AVIR:} A video prefix or complete sequence is paired with auditory evidence or spoken constraints to support continuation or corrective editing.
For each instance, we identify the evidence supporting the target and the semantic requirements that a valid output should satisfy. These requirements concern the inference expressed by the generated video, rather than a unique realization of its appearance or motion.

\paragraph{Implicit Prompt Construction.}
Given the selected observations and semantic target, we construction a prompt that specifies the task while withholding the information to be inferred.
We employ ChatGPT~\citep{openai2022chatgpt} and the open-source Qwen3 model~\citep{yang2025qwen3} to produce candidate formulations and linguistic variations, which are subsequently edited and verified by experts.
We demonstrate that this prompt construction follows two criteria. First, the text alone should not disclose the target interpretation or generation.
Second, the multimodal inputs should provide sufficient evidence to infer the generative results.
\textit{For example}, for AVIR tasks, audio input may specify a constraint, while identifying its violation and determining the required correction remain grounded in the video. 
So, we revise prompts that reveal the answer, obscure the requested generative content, or require assumptions unsupported by the inputs.

\paragraph{Iterative Expert Review.}
Finally, each paired condition-prompt candidate undergoes \textbf{10-loop} expert review along four dimensions: 1) Scene complexity: The scene contains sufficient task-relevant objects, relations, or dynamics to support the intended evaluation. 2) Inferential richness: Satisfying the target requires spatiotemporal, physical, or semantic inference beyond directly reproducing the prompt. 3) Condition alignment: The inputs jointly support the annotated target, with any intentional discrepancy between the video and audio constraints defining the required correction in generations. 4) Prompt implicitness: The prompt leaves the target inference unstated while clearly specifying the task.
Expert reviewers further verify that each condition-prompt pair provides sufficient evidence for the intended inference while allowing plausible variation in the generated video. Pairs that fail review are revised by adjusting the prompt or replacing the input conditions and then reassessed. 

\begin{wrapfigure}{r}{0.48\columnwidth}
    \vspace{-0.8em}
    \centering
    \includegraphics[width=1.0\linewidth]{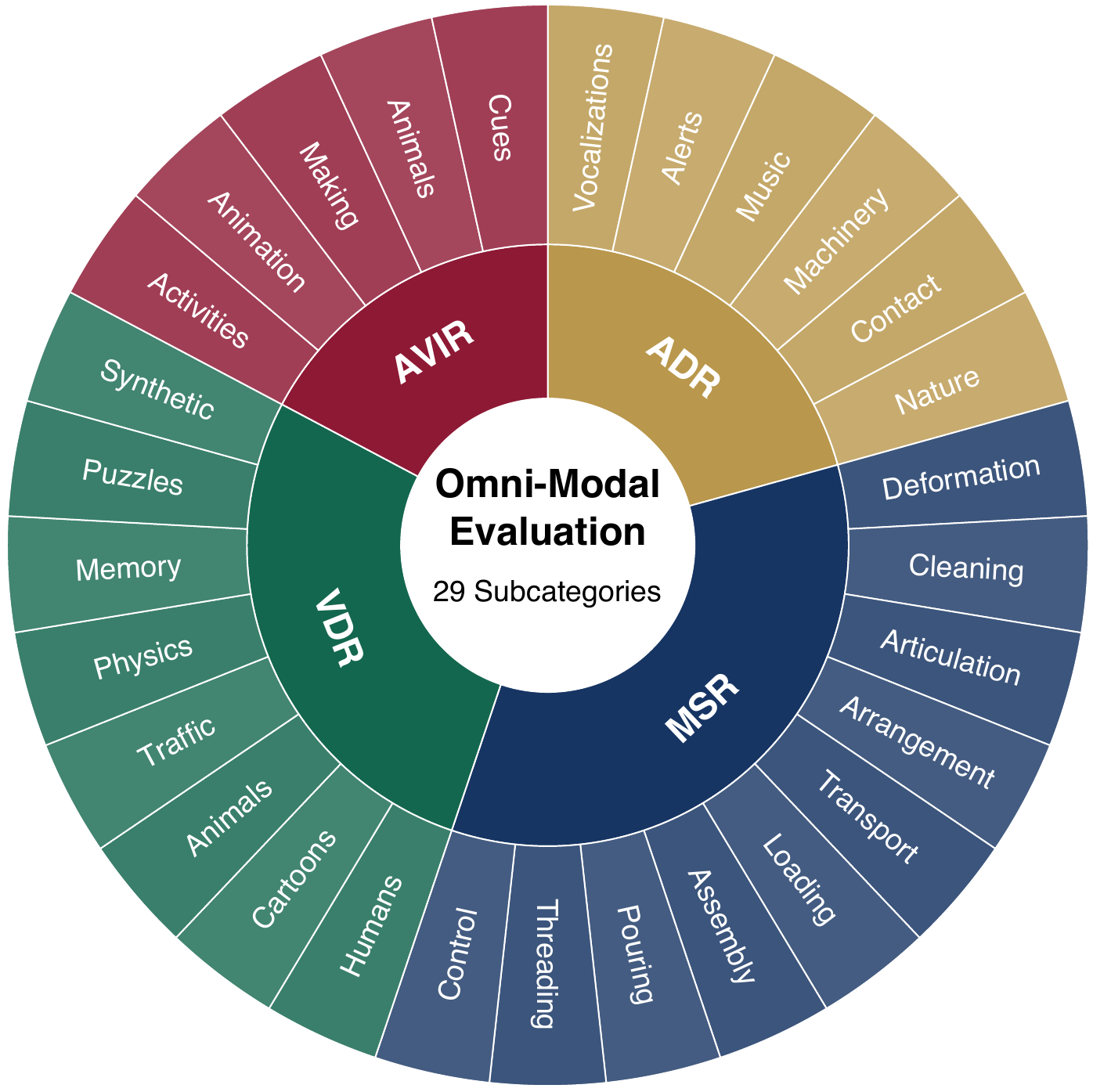}
    \caption{Our evaluation set consists of 4 domains and 29 subdomains, totaling 517 paired conditions.}
    \label{fig:category}
    \vspace{-0.8em}
\end{wrapfigure}

\noindent
After data collection, filtering, and human verification, each accepted instance is represented as
$\mathcal{E}_i = (q_i, \mathbf{x}_i, \tau_i)$,
where $q_i$ denotes the implicit task prompt, $\mathbf{x}_i$ contains the input observations, and
$\tau_i \in \{\mathrm{MSR}, \mathrm{ADR}, \mathrm{VDR}, \mathrm{AVIR}\}$
identifies the reasoning scenario. Depending on the scenario, the observations consist of multiple images, an image paired with audio, a video prefix, or a prefix or complete video paired with audio. Each condition-prompt pair is verified to support the intended inference through video generation for \modelname.
As shown in Fig.~\ref{fig:category}, our evaluation set contains 4 domains and 29 subdomains, totaling 517 paired conditions.

\section{Experiments} \label{sec:experiments}

We evaluate whether \modelname can generate videos that follow the spatial, temporal, and audiovisual constraints implied by multimodal inputs. Our analysis focuses on three aspects: overall success across the four reasoning scenarios, performance differences between task categories, and representative cases in the generated videos.

\subsection{Evaluation Data and Protocol}
\label{sec:data_sources_metrics}

\begin{figure}[t!]
    \centering
    \includegraphics[width=\linewidth]{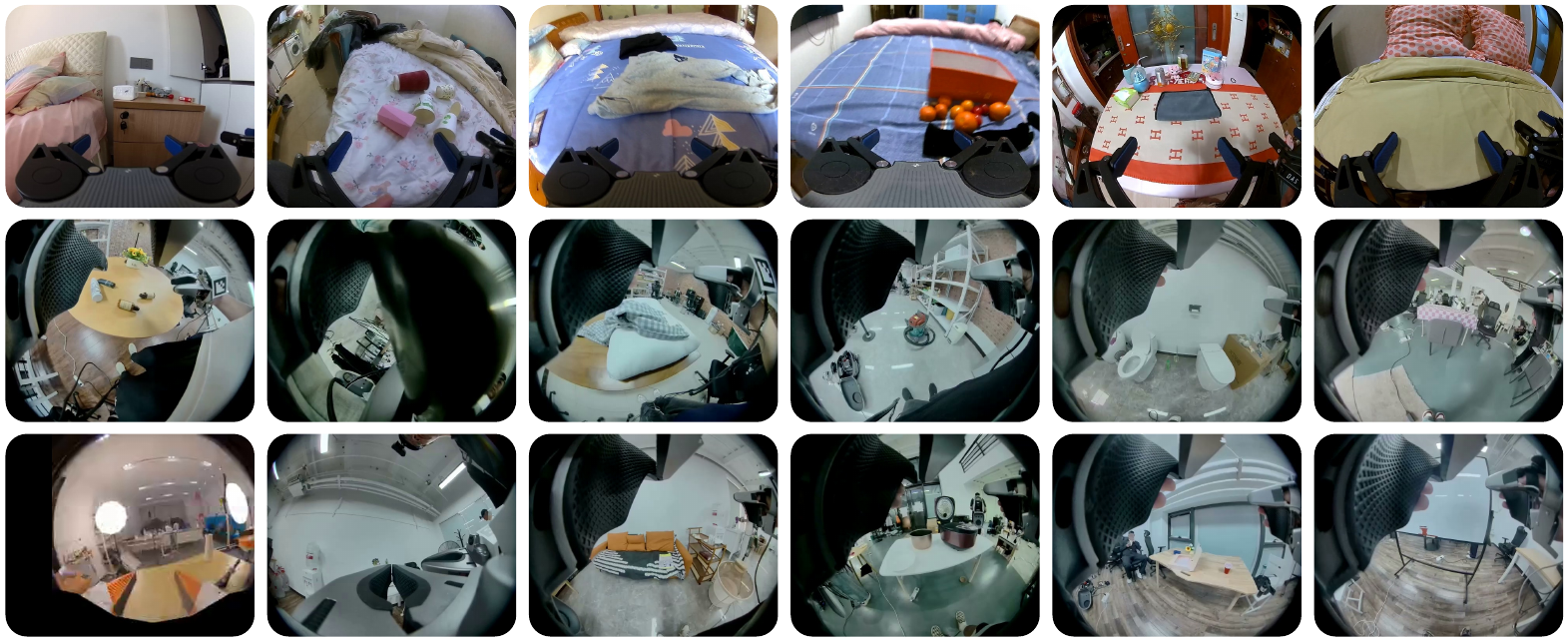}
    \caption{\textbf{MSR evaluation inputs.} Household and tabletop manipulation scenes include substantial viewpoint variation, occlusion, and diverse object configurations.}
    \label{fig:msr_data}
\end{figure}

\begin{figure}[ht]
    \centering
    \includegraphics[width=\linewidth]{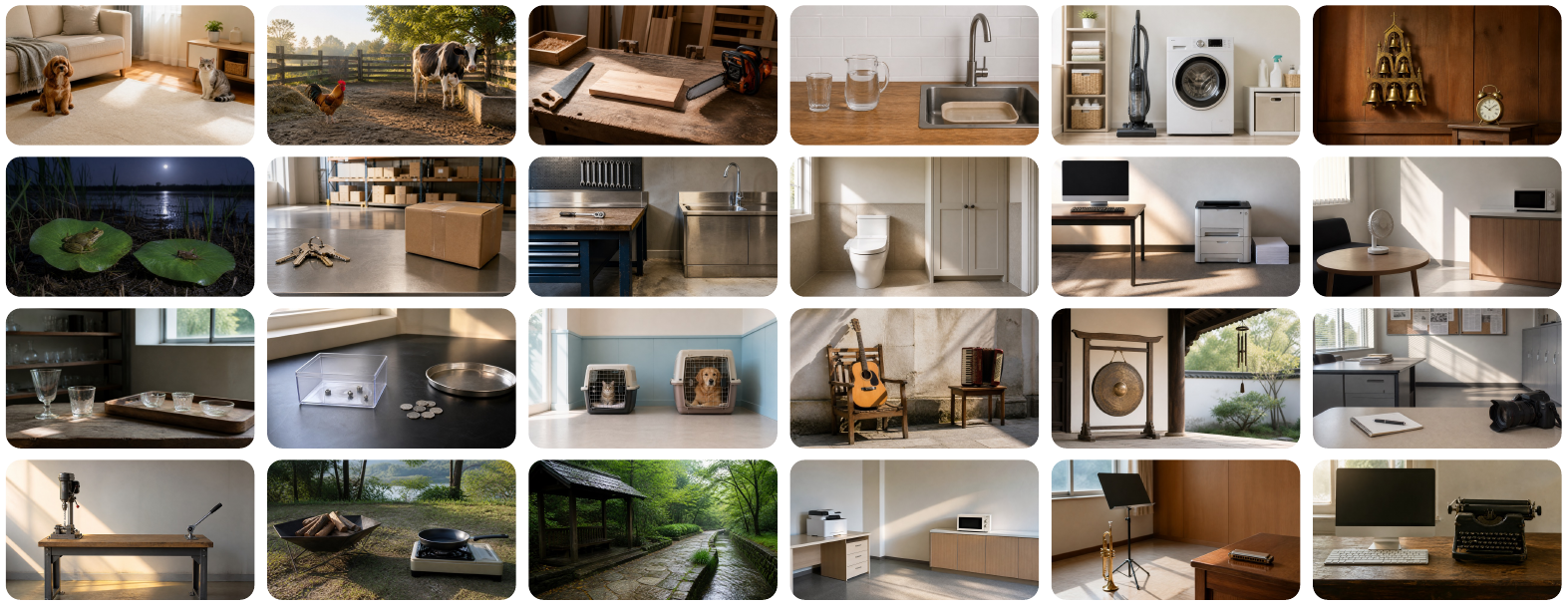}
    \caption{\textbf{ADR evaluation inputs.} Scenes contain alternative candidate sound sources, including animals, instruments, tools, and appliances.}
    \label{fig:adr_data}
\end{figure}

\begin{figure}[t!]
    \centering
    \includegraphics[width=\linewidth]{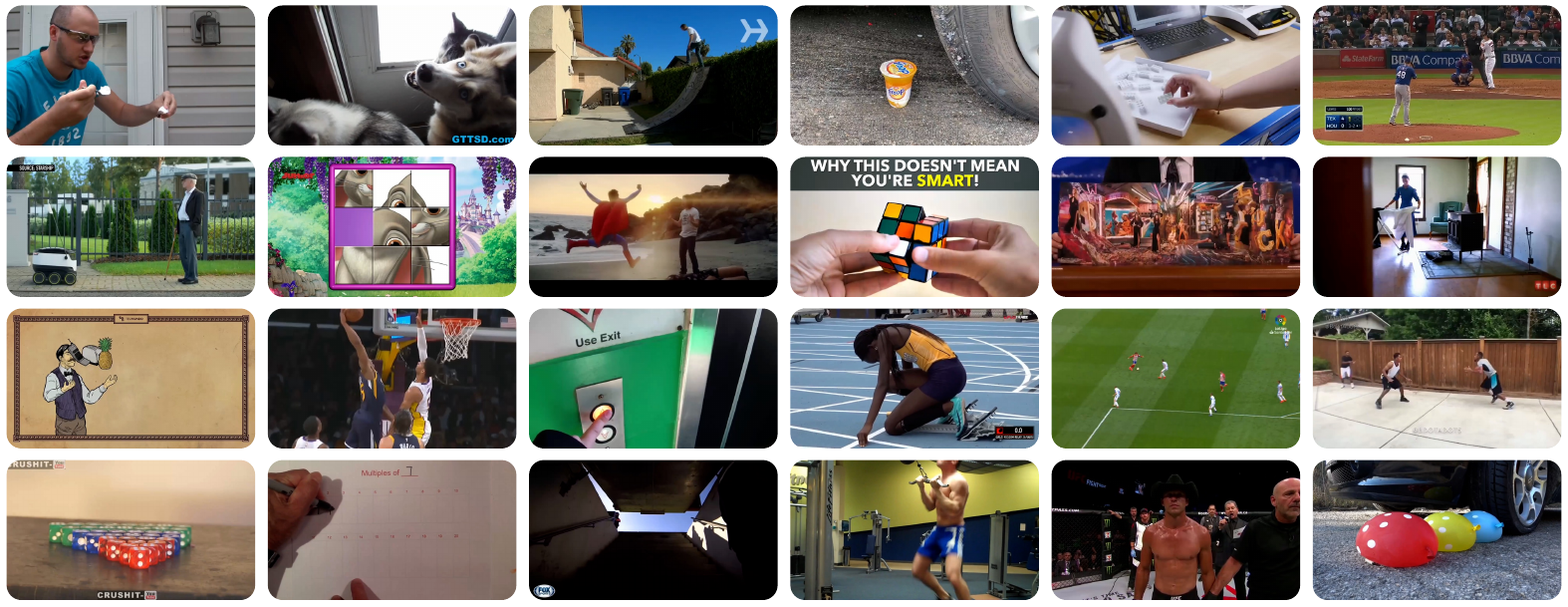}
    \caption{\textbf{VDR evaluation inputs.} Selected video frames illustrate human and animal behavior, physical interactions, puzzles, and scene-memory settings.}
    \label{fig:vdr_data}
\end{figure}

\begin{figure}[tp]
    \centering
    \includegraphics[width=\linewidth]{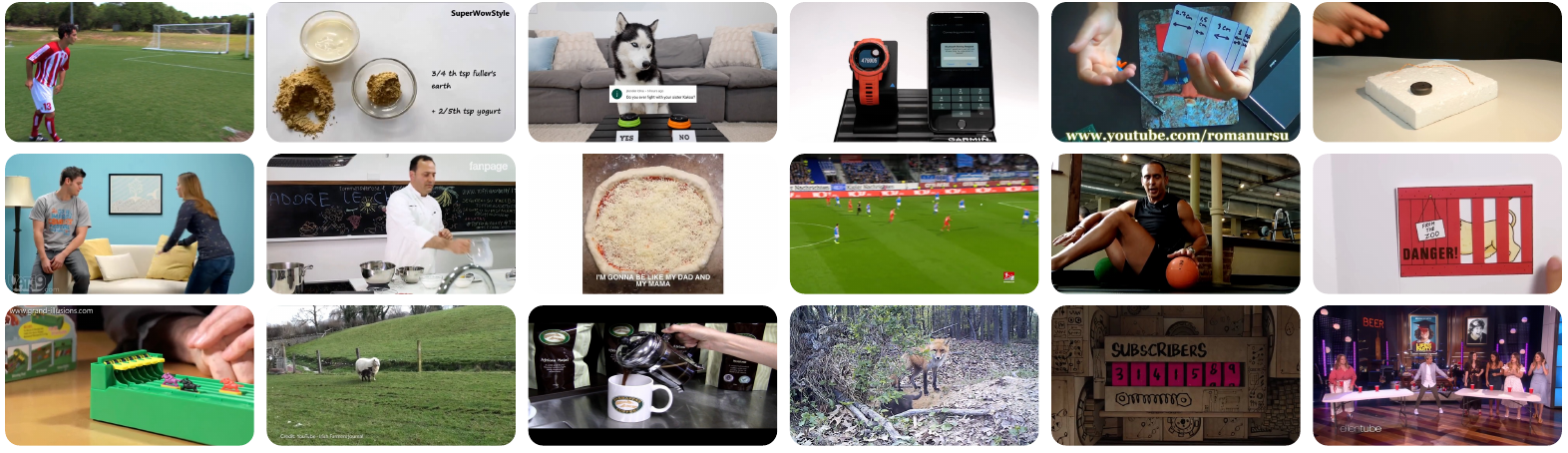}
    \caption{\textbf{AVIR evaluation inputs.} Selected frames span activities, animation, animals, and cues. Video is paired with environmental sounds or spoken constraints for continuation or localization.}
    \label{fig:avir_data}
\end{figure}

\paragraph{Data sources and coverage.}
Our evaluation set contains 517 instances spanning four scenarios and 29 subcategories: MSR (200 instances; 10 subcategories), VDR (100; 8), ADR (146; 6), and AVIR (71; 5). Multi-view visual observations are derived from HiFi-UMI-2K~\citep{ai2026hifi}, VISTA-UMI-5K~\citep{yang2026vista}, HuMI-Unsheathe~\citep{nai2026humanoid}, Hy-Embodied-0.5-VLA-Data~\citep{zhang2026hy}, and 10Kh-RealOmin-OpenData~\citep{genrobot2025data}. Video sources include LLaVA-Video-178K~\citep{zhang2024video}, and acoustic sources include FSD50K~\citep{fonseca2021fsd50k} and ESC-50~\citep{piczak2015esc}. These datasets complement the synthetic sources described in Sec.~\ref{sec:data_construction}. All accepted instances undergo the same input-prompt construction and expert verification.
Figs.~\ref{fig:msr_data}-\ref{fig:avir_data} show representative examples from the four evaluation scenarios.
MSR focuses on household manipulation with multi-view observations, including large viewpoint changes and partial occlusions.
ADR contains scenes with multiple objects that may produce different sounds, requiring the model to use audio cues to identify the corresponding event.
VDR evaluates video continuation across human activities, animal motion, physical interactions, scene changes, and object dynamics.
AVIR combines video with environmental audio or spoken instructions to evaluate audio-guided continuation and visual inconsistency detection.

\paragraph{Human evaluation.}
Three experts independently evaluate each generated video and then cross-check their judgments using the corresponding inputs and task instructions. Since each sample may admit multiple valid outputs, evaluation is based on whether the generated video satisfies the intended task requirement rather than matching a single reference video.
For MSR, reviewers check whether the generated video preserves the required spatial relations and coordinated actions across views. For ADR, they verify whether the generated event is consistent with the provided audio. For VDR, they evaluate whether the continuation follows the observed motion and task constraints. For AVIR, they assess whether the audio is correctly reflected in the relevant visual content.
A visually plausible video is therefore considered unsuccessful if it does not satisfy the required task condition.

\paragraph{Metric and scope.}
We use success rate (SR) as the main evaluation metric. For each task category, SR is computed as the percentage of samples that are judged successful according to the corresponding task-specific criteria. Given a subset $\mathcal{D}$ with binary labels $s_i \in \{0,1\}$, we compute:
\begin{equation}
\mathrm{SR}(\mathcal{D}) =
\frac{100}{|\mathcal{D}|}
\sum_{i \in \mathcal{D}} s_i.
\label{eq:exp_success_rate}
\end{equation}

The overall SR is computed over all evaluated samples. We report results for \modelname as a representative omni-modal model, and use these experiments to analyze model performance across different physical-world reasoning scenarios and input conditions.

\subsection{Quantitative Results}
\label{sec:quantitative_results}

\begin{table*}[t]
\centering
\caption{
\textbf{Human-evaluated performance of \modelname across four reasoning scenarios.}
Scenario-level and overall success rates are weighted by sample count.
$N$: number of samples.
SR: success rate ($\uparrow$).
Blank cells indicate no additional subcategories.
}
\label{tab:scenario_results}
\footnotesize
\setlength{\tabcolsep}{3pt}
\renewcommand{\arraystretch}{1.15}

\begin{tabular*}{\textwidth}{
@{\extracolsep{\fill}}
lrr lrr lrr lrr
@{}
}
\toprule
\multicolumn{3}{c}{\textbf{MSR}} &
\multicolumn{3}{c}{\textbf{VDR}} &
\multicolumn{3}{c}{\textbf{ADR}} &
\multicolumn{3}{c}{\textbf{AVIR}} \\
\cmidrule(lr){1-3}
\cmidrule(lr){4-6}
\cmidrule(lr){7-9}
\cmidrule(lr){10-12}

Subcategory & $N$ & SR (\%) &
Subcategory & $N$ & SR (\%) &
Subcategory & $N$ & SR (\%) &
Subcategory & $N$ & SR (\%) \\
\midrule

Deformation   & 34 & 41.20 &
Humans        & 34 & 55.88 &
Vocalizations & 16 & 31.30 &
Activities    & 14 & 57.14 \\

Cleaning      & 27 & 33.30 &
Cartoons      &  5 & 20.00 &
Alerts        & 26 & 23.10 &
Animation     &  6 & 33.33 \\

Articulation  & 26 & 34.60 &
Animals       & 11 & 63.64 &
Music         & 16 & 25.00 &
Making        & 21 & 52.38 \\

Arrangement   & 23 & 47.80 &
Traffic       &  9 & 100.00 &
Machinery     & 34 & 17.60 &
Animals       & 10 & 30.00 \\

Transport     & 23 & 34.80 &
Physics       & 10 & 50.00 &
Contact       & 32 & 21.90 &
Cues          & 20 & 50.00 \\

Loading       & 20 & 50.00 &
Memory        &  6 & 66.67 &
Nature        & 22 & 54.50 &
              &    &       \\

Assembly      & 20 & 50.00 &
Puzzles       &  6 & 16.67 &
              &    &       &
              &    &       \\

Pouring       & 12 & 58.30 &
Synthetic     & 19 & 52.63 &
              &    &       &
              &    &       \\

Threading     &  8 & 75.00 &
              &    &       &
              &    &       &
              &    &       \\

Control       &  7 & 42.90 &
              &    &       &
              &    &       &
              &    &       \\

\midrule
\textbf{Overall} & \textbf{200} & \textbf{43.50} &
\textbf{Overall} & \textbf{100} & \textbf{56.00} &
\textbf{Overall} & \textbf{146} & \textbf{27.40} &
\textbf{Overall} & \textbf{71}  & \textbf{47.89} \\
\bottomrule
\end{tabular*}

\begin{minipage}{\textwidth}
\vspace{3pt}
\footnotesize
Overall success rate across all 517 samples: 41.97\%.
\end{minipage}
\end{table*}

\paragraph{Overall reliability remains limited.}
Table~\ref{tab:scenario_results} reports an overall SR of 41.97\% across 517 instances for \modelname. Among the four scenarios, VDR achieves the highest SR (56.00\%), followed by AVIR (47.89\%), MSR (43.50\%), and ADR (27.40\%). These results show that the model still fails on more than half of the evaluated cases, despite supporting all required input modalities.
The largest performance gap is observed between VDR and ADR, with a difference of 28.60 percentage points. This suggests that the model handles video-based continuation more reliably than audio-dependent reasoning on our test data. However, since the scenarios differ in data, prompts, and generation targets, the results mainly reflect task-level performance rather than a direct comparison between video and audio modalities.

\paragraph{Results across scenarios.}
Furthermore, in Table~\ref{tab:scenario_results}, we analyze that MSR reaches an SR of 43.50\%. Performance varies notably across manipulation tasks: threading achieves 75.00\% and pouring 58.30\%, while cleaning, articulation, and transport remain around 33-35\%. This suggests that maintaining spatial consistency across views is still difficult for several manipulation settings.
VDR performs best overall, reaching 56.00\% SR. The model handles human, animal, and traffic dynamics relatively well, but performance drops substantially on cartoons and puzzles, where success rates fall to 20.00\% and 16.67\%, respectively.
ADR is the most challenging scenario, with an SR of only 27.40\%. Nature sounds are handled better than other categories, whereas machinery, contact, and alert sounds remain difficult. This highlights the challenge of converting acoustic evidence into the correct visual event.
Finally, AVIR reaches 47.89\% SR. The model performs better on activities and making tasks, while animation and animal-related cases remain more difficult. Overall, the results show clear differences across reasoning scenarios and substantial room for improvement.

\begin{figure}[!t]
    \centering
    \includegraphics[width=\linewidth,height=0.78\textheight,keepaspectratio]{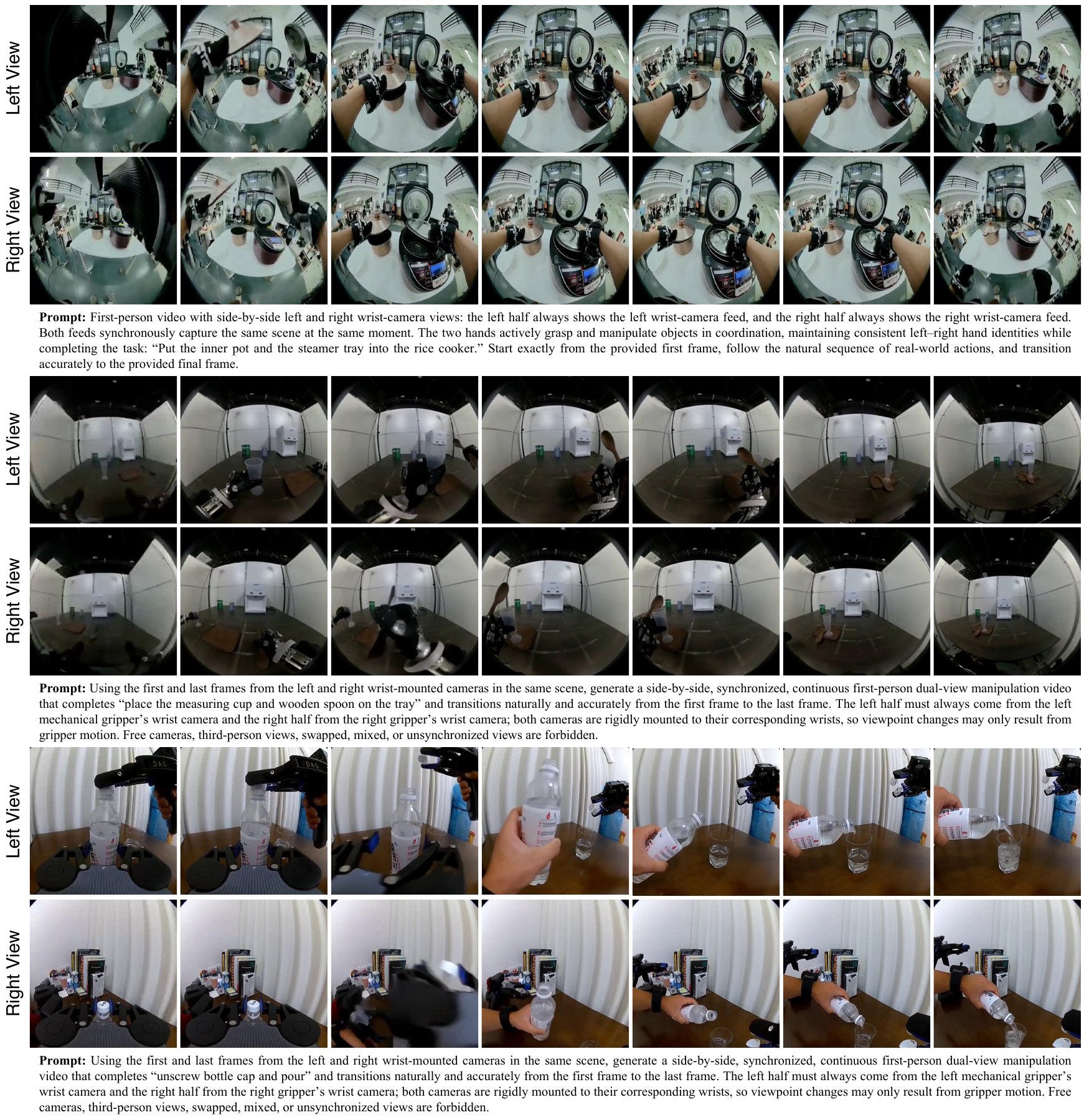}
    \caption{\textbf{MSR results under paired-view conditioning.} Each example shows the paired views together with the corresponding prompt.}
    \label{fig:msr_results}
\end{figure}

\subsection{Qualitative Analysis}
\label{sec:qualitative_results}

\begin{figure}[tp]
    \centering
    \includegraphics[width=\linewidth,height=0.78\textheight,keepaspectratio]{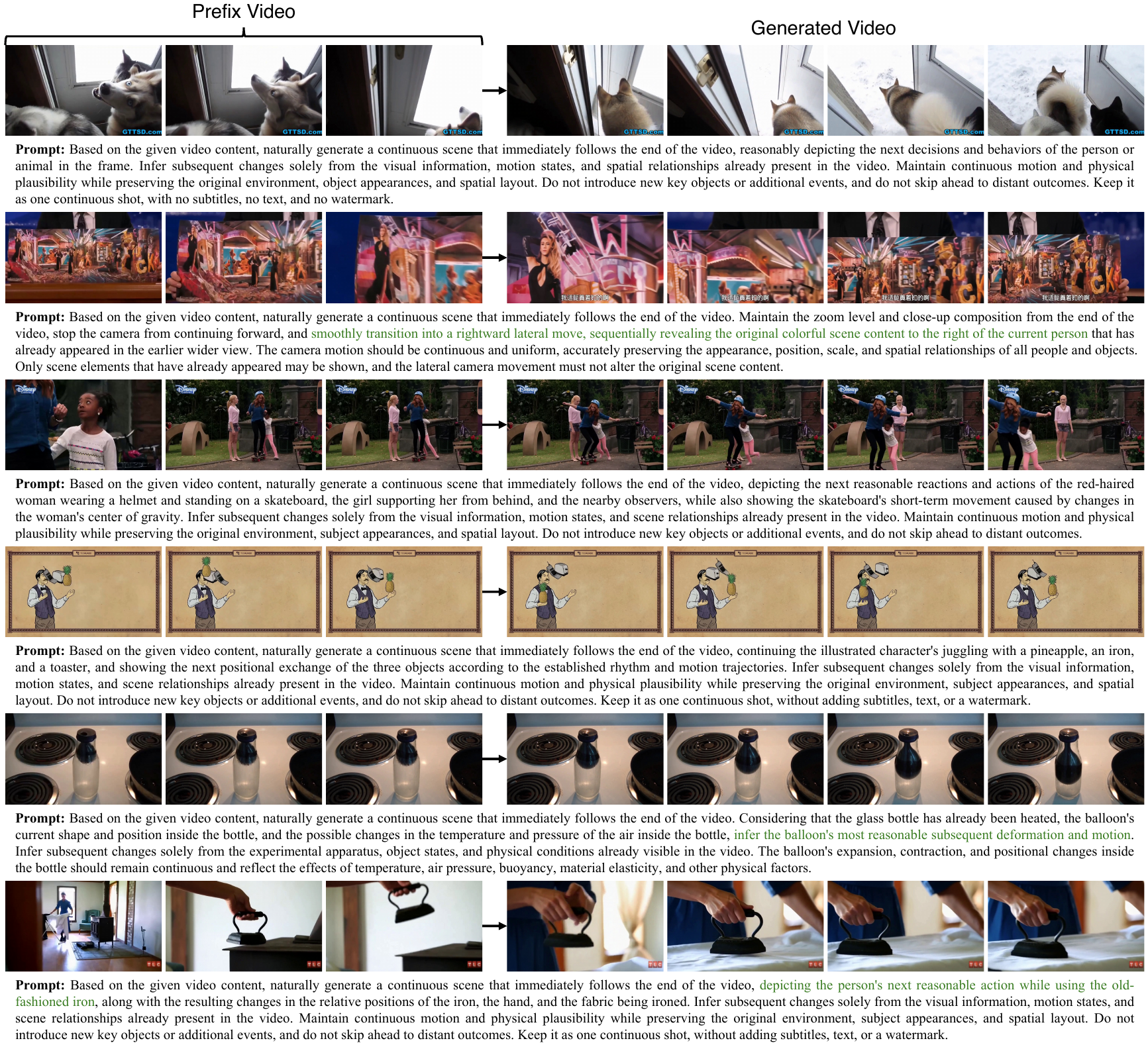}
    \caption{\textbf{VDR results conditioned on prefix videos.} Each row shows observed frames on the left and generated continuation frames on the right, together with the corresponding instruction.}
    \label{fig:vdr_results}
\end{figure}

\begin{figure}[tp]
    \centering
    \includegraphics[width=\linewidth,height=0.78\textheight,keepaspectratio]{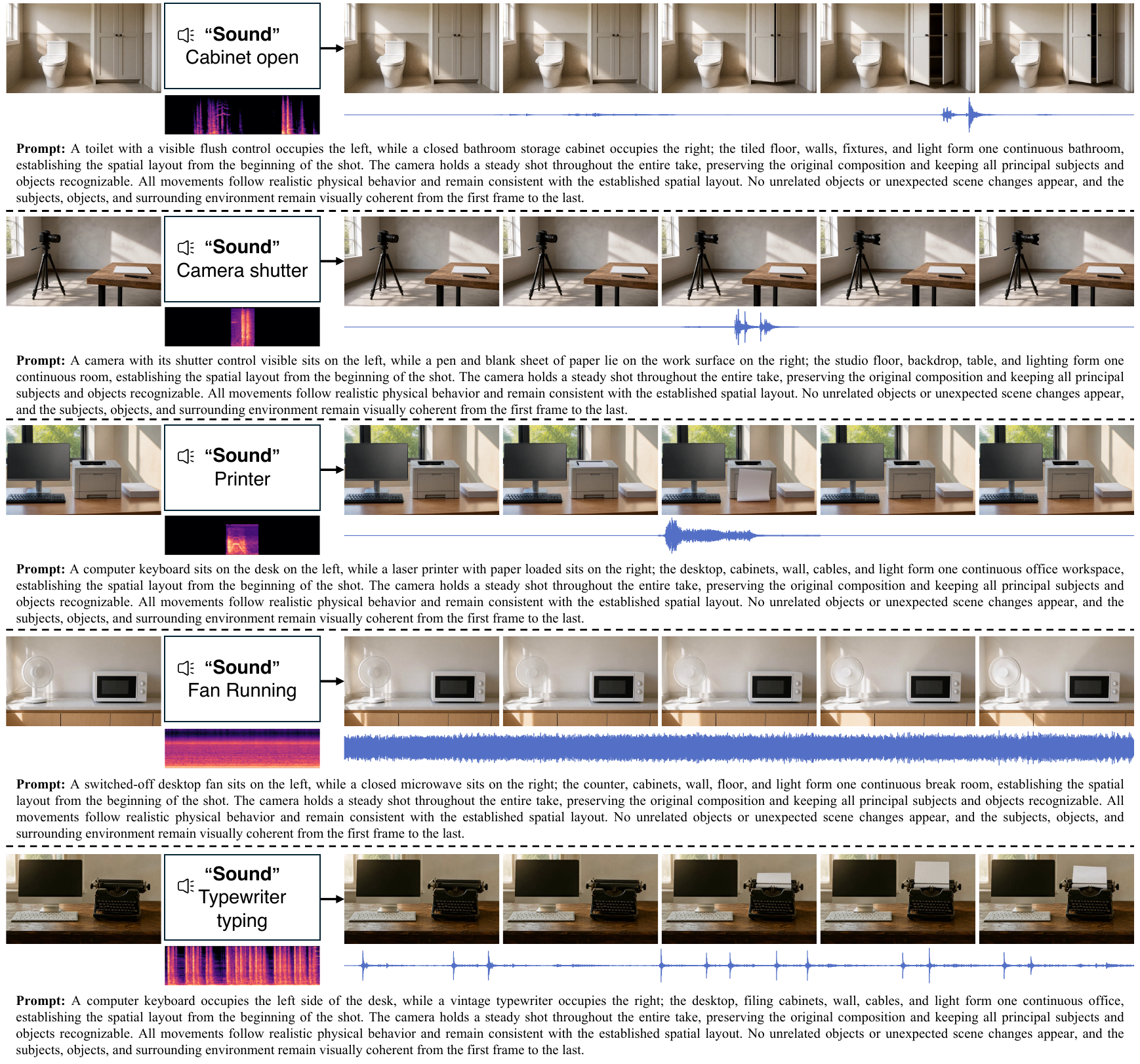}
    \caption{\textbf{ADR results conditioned on visual and acoustic inputs.} Each row shows the input scene, acoustic cue, and generated frames. The cabinet-opening and typewriter examples produce visible events consistent with the supplied sounds.}
    \label{fig:adr_results}
\end{figure}

\begin{figure}[tp]
    \centering
    \includegraphics[width=\linewidth,height=0.78\textheight,keepaspectratio]{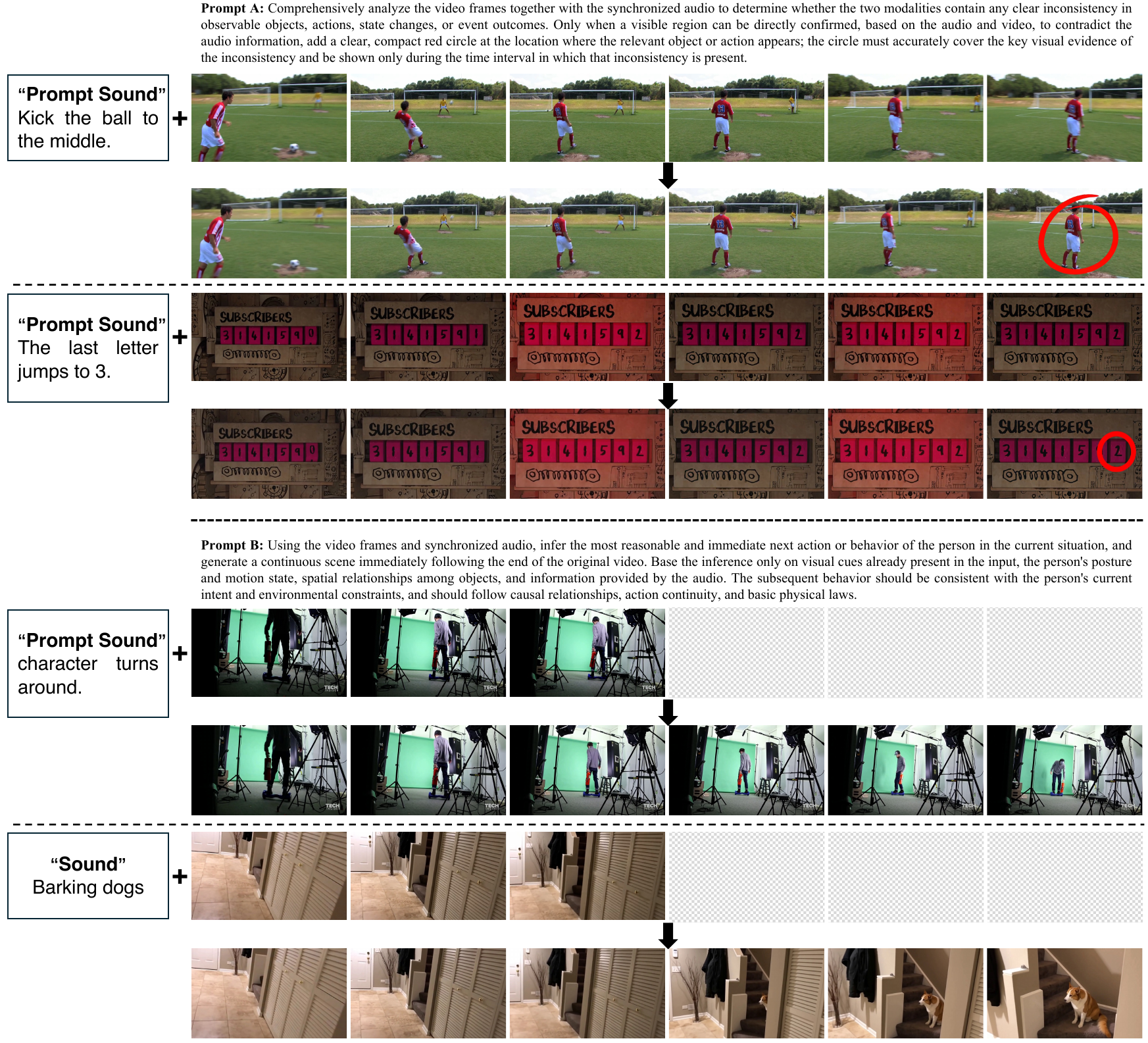}
    \caption{\textbf{AVIR results for discrepancy localization and audiovisual continuation.} The upper examples localize visual content that conflicts with spoken constraints, while the lower examples continue the video from audiovisual inputs.}
    \label{fig:avir_results}
\end{figure}

Figs~\ref{fig:msr_results}-\ref{fig:avir_results} present representative qualitative results across the four evaluation scenarios. Overall, the model can generate visually plausible continuations and often follows the dominant cues provided by the input modalities. In MSR, it produces recognizable manipulation stages for tasks such as rice-cooker assembly and utensil placement, while in VDR it generates reasonable short-term dynamics, such as a dog approaching a doorway or an iron contacting fabric. ADR and AVIR further show that audio can guide the generated visual content: the model opens the cabinet in response to the corresponding sound, produces paper motion for typewriter audio, and follows several spoken or environmental cues during continuation.

\subsection{Failure Cases}
\label{sec:failure_cases}

\begin{figure}[tp]
    \centering
    \includegraphics[width=\linewidth,height=0.78\textheight,keepaspectratio]{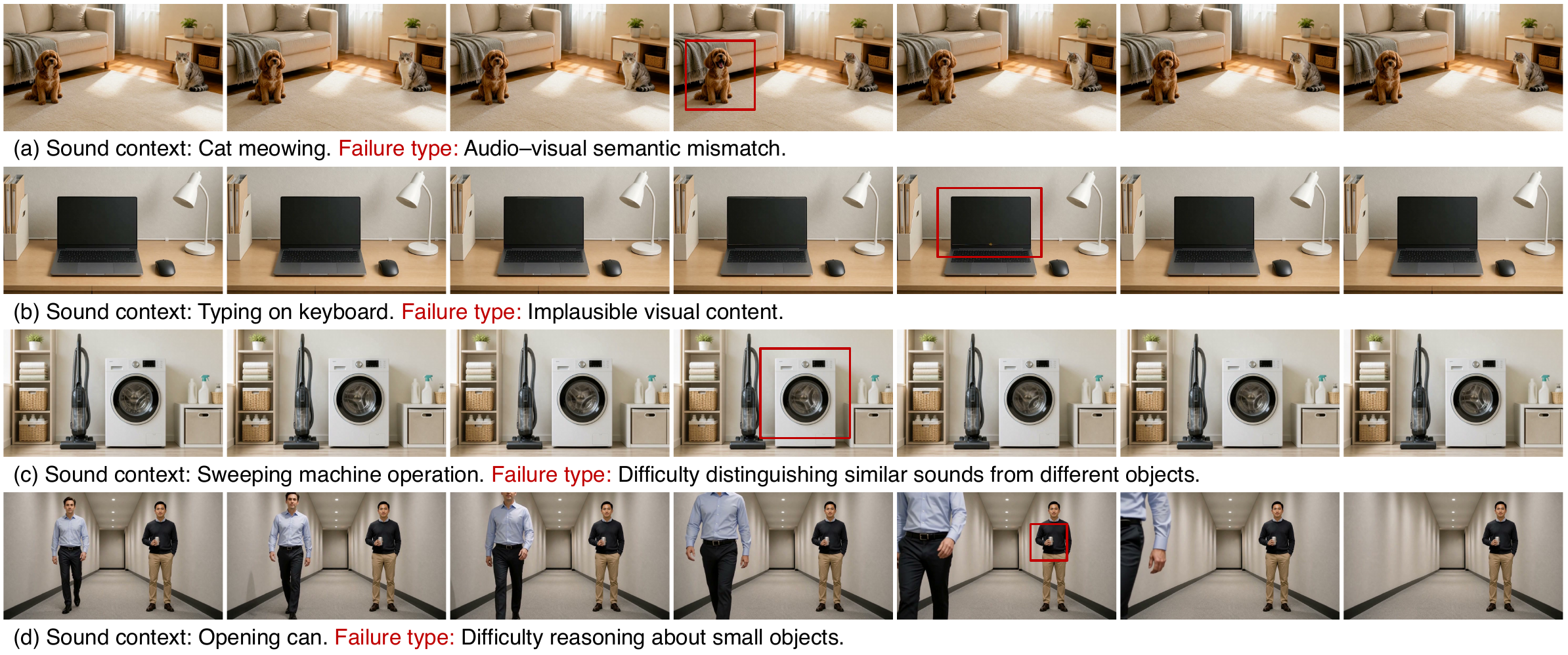}
    \caption{\textbf{Representative ADR failure cases.} Each row shows uniformly sampled video frames and identifies the input sound and failure type: (a) audio-visual semantic mismatch, (b) implausible visual content, (c) confusion between similar sound sources, and (d) failure to ground a small-object interaction. Red boxes highlight the regions relevant to each failure.}
    \label{fig:adr_failure}
\end{figure}

\begin{figure}[tp]
    \centering
    \includegraphics[width=\linewidth,height=0.78\textheight,keepaspectratio]{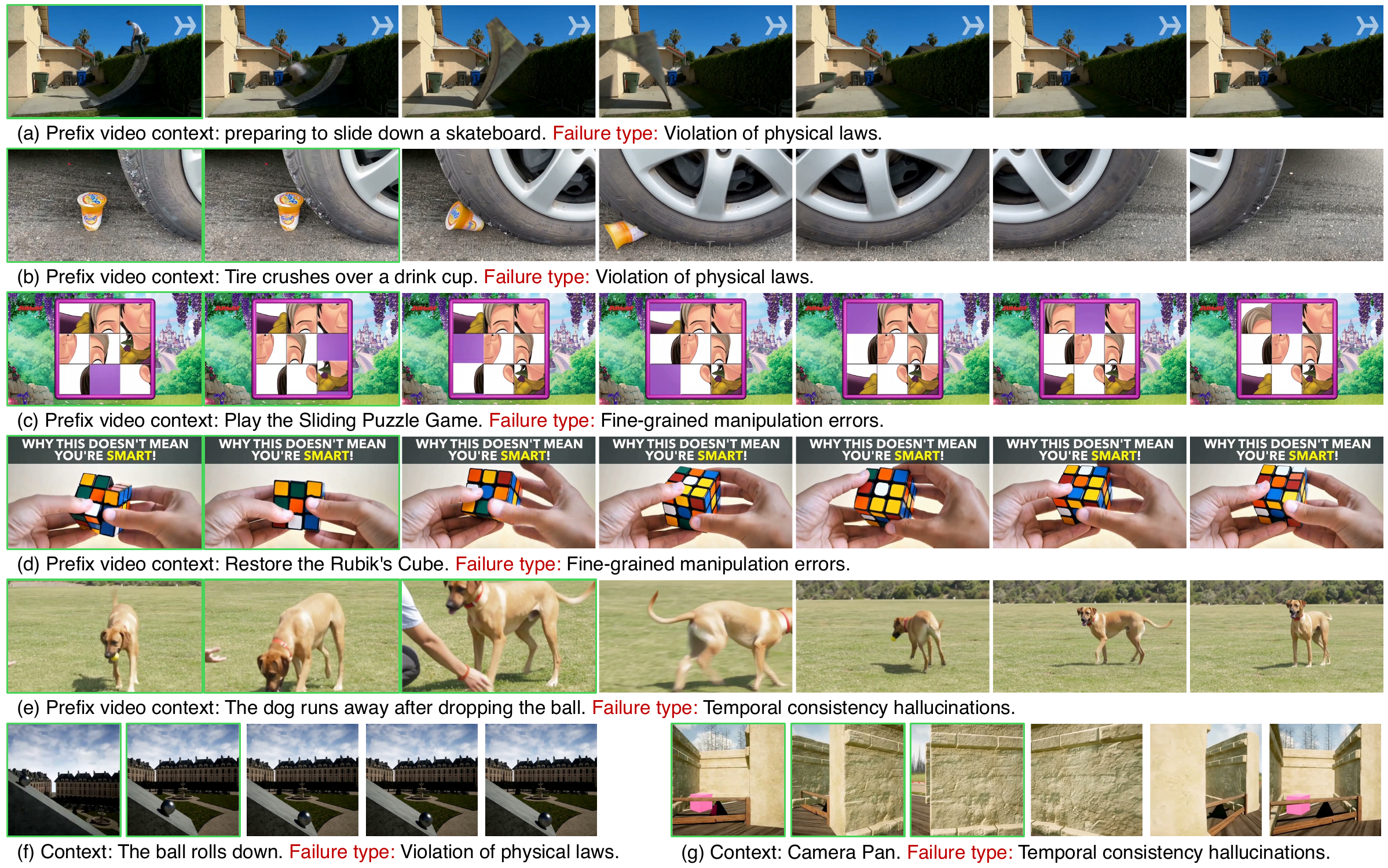}
    \caption{\textbf{Representative VDR failure cases.} Green-bordered frames indicate the observed prefix, followed by sampled continuation frames. Panels (a), (b), and (f) illustrate violations of physical constraints; (c) and (d) illustrate fine-grained manipulation errors; and (e) and (g) illustrate temporal inconsistencies in object states and scene content.}
    \label{fig:vdr_failure}
\end{figure}

Moreover, we also analyze the failure cases of the tested results.
Figs.~\ref{fig:adr_failure} and~\ref{fig:vdr_failure} reveal several recurring failure patterns across multimodal physical-world reasoning tasks. We list the failure types for audio and video inputs: \textbf{1)} Incorrect evidence grounding: occurs when the model identifies a plausible event but associates the conditioning evidence with the wrong entity or action, as in the cat-dog and cleaning-appliance examples. \textbf{2)} Incomplete event realization: appears when the generated scene contains relevant objects but fails to instantiate the interaction implied by the input, such as typing or can opening. \textbf{3)} Physical and configurational violations: arise when generated continuations break contact dynamics, object geometry, or valid state transitions, as observed in the skateboarding, rolling-object, puzzle, and Rubik's-Cube cases. \textbf{4)} temporal state inconsistency: occurs when previously established object states, motion trends, or scene content are not preserved throughout the continuation.
These failures suggest a common challenge beyond perceptual plausibility: the model must convert multimodal evidence into the correct event while preserving the physical, spatial, and temporal constraints established by the observations. 

For Omni-Models, the ability to jointly understand and reason across multiple modalities is essential. Generative Omni-Modal models further make this reasoning process observable through visual generation, where intermediate predictions can be interpreted as a form of \emph{Chain-of-Frames}. The failure cases identified above therefore provide a concrete view of where current models still fall short, and suggest clearer directions for improving cross-modal grounding, physical reasoning, and temporally consistent generation in future Omni-Modal systems.

\subsection{Discussion}
\label{sec:experimental_insights}

Our experiments provide three main observations about physical-world reasoning in omni-modal generative models, particularly \modelname.
\textbf{1) Visual plausibility does not guarantee physical-world consistency.}
The model can often generate realistic videos, yet still violate important spatial, temporal, and audiovisual constraints. Typical failures include inconsistent embodiment across views, incorrect camera motion, incomplete state transitions, missing visual responses to audio cues, and imprecise audiovisual localization.
\textbf{2) Multimodal support does not necessarily lead to effective multimodal reasoning.}
Although the model accepts images, videos, audio, and text as inputs, our results show that it does not always use these signals reliably to satisfy the task requirements. This gap is particularly evident in tasks that require acoustic grounding or coordination across multiple views.
\textbf{3) Generation-based evaluation reflects the complete reasoning-and-generation process.}
A failed output may arise from incorrect input understanding, weak cross-modal integration, or errors during video generation. Further controlled experiments, such as removing individual modalities or replacing audio while keeping the visual input fixed, could help separate these factors.

\section{Conclusion}
\label{sec:applications}

In this work, we investigated physical-world reasoning in emerging Omni-Modal Generative Models through the lens of multimodal generation. Rather than evaluating generation under fully specified prompts, we constructed implicit condition--prompt pairs in which critical information must be recovered from complementary evidence distributed across multiple modalities. This formulation enables us to examine whether an Omni-Model can move beyond accepting heterogeneous inputs and effectively integrate them to infer latent event states, physical dynamics, and appropriate outcomes.
Our evaluation of \modelname demonstrates both the promise and the current limitations of this capability. While complementary multimodal evidence can support reasoning beyond explicitly stated instructions, the overall performance remains limited, revealing a substantial gap between omni-modal input support and effective physical-world reasoning. More broadly, our results suggest that omni-modal inputs provide not only a richer interface for content generation, but also a useful foundation for constructing new evaluation paradigms that probe reasoning through generation.

Looking forward, we plan to continuously expand and refine the evaluation set with more diverse physical-world scenarios, modality combinations, and reasoning requirements. We also aim to develop an automated evaluation framework that can reliably assess reasoning outcomes in generated content, enabling scalable and reproducible testing of future Omni-Models.

\newpage
\bibliographystyle{plainnat}
\bibliography{neurips_2026}

\end{document}